\pdfoutput=1

\documentclass[11pt]{article}

\usepackage[]{ACL2023}
\usepackage{xcolor}
\usepackage{times}
\usepackage{latexsym}
\usepackage{booktabs}
\usepackage{enumitem}

\usepackage[T1]{fontenc}
\usepackage{graphicx}
\usepackage[utf8]{inputenc}

\usepackage{microtype}

\usepackage{inconsolata}

\usepackage{soul}
\usepackage{multirow}
\usepackage{cleveref}
\title{CQSumDP: A {C}hatGPT-Annotated Resource for {Q}uery-Focused Abstractive {Sum}marization Based on {D}ebate{p}edia}

\author{Md Tahmid Rahman Laskar\textsuperscript{1,3}, \  Mizanur Rahman\textsuperscript{2,3}, \  Israt Jahan\textsuperscript{3}, \\ {\bf Enamul Hoque\textsuperscript{3}, \ Jimmy Huang\textsuperscript{3,}\thanks{\hspace{0.115cm}Contact Author.}}\\
       \textsuperscript{1}Dialpad Canada Inc.,  \textsuperscript{2}Royal Bank of Canada,  \textsuperscript{3}York University\thanks{\hspace{0.115cm}All work being done at York University.} \\ Toronto, Ontario, Canada\\
       \texttt{\{tahmid20,mizanurr,israt18,enamulh,jhuang\}@yorku.ca}}

\begin{document}
\maketitle

\begin{abstract}
 Debatepedia is a publicly available dataset consisting of arguments and counter-arguments on controversial topics that has been widely used for the single-document query-focused abstractive summarization task in recent years. However, it has been recently found that this dataset is limited by noise and even most queries in this dataset do not have any relevance to the respective document. In this paper, we present a methodology for cleaning the Debatepedia dataset by leveraging the generative power of large language models to make it suitable for query-focused abstractive summarization. More specifically, we harness the language generation capabilities of ChatGPT to regenerate its queries. We evaluate the effectiveness of the proposed ChatGPT annotated version of the Debatepedia dataset using several benchmark summarization models and demonstrate that the newly annotated version of Debatepedia outperforms the original dataset in terms of both query relevance as well as summary generation quality. We will make this annotated and cleaned version of the dataset publicly available. 
\end{abstract}

\section{Introduction}
Abstractive summarization is a natural language processing technique that involves generating a concise and coherent summary of a longer piece of text while preserving its most important information \cite{yao2017survey}. Query-focused abstractive summarization is a specific type of abstractive summarization that generates a summary of the given text that is tailored to a specific query or topic of interest \cite{baumel2018query,goodwin2020flightpegasus,su2020cairecovid,xu2021text,laskar2020query,laskar2020wslds,laskar2022domain}. In other words, the summary is focused on answering a specific question or addressing a particular topic, rather than providing a general overview of the text. One widely used dataset for this task is the Debatepedia\footnote{\url{https://github.com/PrekshaNema25/DiverstiyBasedAttentionMechanism}} dataset that consists of arguments and counter-arguments on conversational topics \cite{nema}.

The query-focused summarization of argumentative text is a challenging task that has gained increasing attention in recent years due to its potential applications in various domains, such as policy-making, journalism, and legal reasoning \cite{nema, laskar2020query}. However, it has been recently found that the quality of the Debatepedia dataset that is widely used for the query-focused abstractive summarization task is limited by noise, with many of the queries in this dataset does not have any relevance with the source document \cite{laskar2022domain}.  Since Debatepedia is a rich source of argumentative text on controversial topics that can serve as a valuable resource for developing and evaluating summarization models, in this paper, we present a novel methodology to annotate the Debatepedia dataset to make it a useful resource for query-focused abstractive summarization. Our data annotation approach leverages the language modeling \cite{gpt2} capabilities of ChatGPT\footnote{\url{https://openai.com/blog/chatgpt/}}, a large pre-trained language model \cite{bert,gpt3,instructgpt} that has shown an impressive capability of generating fluent and coherent text \cite{qin2023chatgpt,bang2023multitaskchatgpt,yang2023exploringchatgpt,kuzman2023chatgpt,gao2023exploring,wang2023chatgpt,zhou2023comprehensive,kocon2023chatgpt,kocmi2023large}. Using ChatGPT as the annotator, we regenerate the queries in the Debatepedia dataset to remove the noise in this dataset. We validate the effectiveness of our methodology by conducting extensive experiments on our newly constructed dataset that leverages ChatGPT as the annotator. Our major contributions in this paper are summarized below: 

\begin{itemize}
    \item We proposed a novel methodology for cleaning and annotation of the Debatepedia dataset using a large language model, i.e., ChatGPT to improve its suitability for query-focused abstractive summarization. This paper also opens up a promising avenue to utilize ChatGPT as the annotator for other tasks beyond text summarization that can significantly reduce the overall cost of data annotation.
    
    \item We conducted extensive experiments using benchmark summarization models on our ChatGPT-annotated cleaned version of Debatepedia for Query-Focused Abstractive Summarization and observe that it outperforms the original dataset in terms of both query relevance and summary generation quality. 

    \item Our annotated dataset will be made publicly available such that it can serve as a valuable resource for further research on query-focused abstractive summarization. 
    
\end{itemize}

\begin{table*}[]
\small
\begin{tabular}{p{15cm}}

\toprule
\textit{\textbf{Example 1:} Query having no relevance with the document and the summary.}  \\ \midrule
\textbf{Query:} Does an MBA enhance leadership skills? \\ \midrule
\textbf{Document:} Business schools might improve your quantitative presentation and communication skills. It might but get you thinking about ethical and strategy. But two years of case studies aren’t go to turn you into a leader if you weren’t died one. There’s no learning charisma persuasiveness elegance or gut instinct.       \\ \midrule 
\textbf{Reference Summary:} PhD will not improve cm factors of leaders.
\\ \midrule 
\midrule
\textit{\textbf{Example 2:} One word summary having no relevance with the query or document.}
 \\ \midrule
\textbf{Query:} Education : do child benefit from watching tv? \\ \midrule
\textbf{Document:} by watching news child can learn about geography politics advances in science -- everything simply and later explained . furthermore child learn about real-life situation that happens on everyday basis which will benefit them in the future. \\ \midrule
\textbf{Reference Summary:} News. 
\\ \midrule 
\midrule
\textit{\textbf{Example 3:} The length of the summary is longer than the document with the query being irrelevant.} 
 \\ \midrule
\textbf{Query:} activists : where do the keys activists and organizations stand ?  \\ \midrule
\textbf{Document:} see an analyses of the article ... \\ \midrule
\textbf{Reference Summary:} philip martin of berkeley davis and michael teitelbaum the mirage of mexican guest workers nov/dec \# foreign affairs .
\\ \midrule 
\midrule
\textit{\textbf{Example 4:} More of a close-ended question.} 
 \\ \midrule
\textbf{Query:} friendships : does twitter harms relationships ?  \\ \midrule
\textbf{Document:} twitter helps those stay in touches no matter how far they may be from each other . \\ \midrule
\textbf{Reference Summary:}  long-distance friendships . 
\\ \bottomrule
\end{tabular}
\caption{Some examples demonstrating the limitations in the Debatepedia dataset.}
\label{tab:q_e_1}
\end{table*}

\section{Related Work}

Query-focused abstractive summarization using neural models has gained increasing attention in recent years \cite{baumel2018query, laskar2022domain}. The recent success of transformer-based encoder-decoder models \cite{liuemnlpbertsum,bart,t5,zhang2019pegasus} on generic\footnote{In Generic Abstractive Summarization, the summaries are generated based on only the given source document.} abstractive summarization has also inspired researchers to utilize such models for query-based abstractive summarization \cite{goodwin2020flightpegasus,vig2021exploring,laskar2020query,laskar2020wslds,laskar2022domain}, leading to state-of-the-art performance in benchmark query-based summarization and answer generation datasets, such as DUC\footnote{\url{https://duc.nist.gov/data.html}} \cite{queryfocusedsummarization2017unsupervised, dualces, xu2021text,xulapata2020coarseemnlp}, AQuaMuSe \cite{kulkarni2020aquamusegoogle}, QMSum \cite{zhong2021qmsum}, WikiHowQA \cite{deng2020jointanswerselectionsummarygeneration}, PubMedQA \cite{jin2019pubmedqa}, MediQA \cite{savery2020questionmediqaans}, MS-MARCO \cite{msmarco}, Debatepedia \cite{nema}, etc. Though some studies \cite{abdullah2020towardsinlg} also attempted to generate the queries in generic summarization datasets (e.g., CNNDM \cite{nallapati}) using the source document and the reference summary to enable such datasets for query-focused summarization, we find that these queries are generated by directly extracting words or tokens from the reference summaries. As a result, the summarization models have unexpected access to the keywords in the gold reference summaries.

Among the datasets mentioned above, DUC and AQuaMuSe require generating summaries from multiple documents, usually from the news domain. The QMSum dataset is proposed for query-based meeting summarization, while WikiHowQA is constructed from the WikiHow knowledgebase and used for answer summary generation for questions that start with ``How to''. Meanwhile, PubMedQA and MediQA datasets are constructed from the biomedical domain.  One notable exception among these datasets is the Debatepedia dataset since it requires generating abstractive summaries from a short document containing argumentative text. None of the other datasets mentioned above addressed the issue of generating query-based summaries from documents containing arguments and counter-arguments. This makes Debatepedia a great resource for researchers to develop methods to summarize a short document containing argumentative text for the given query. 

However, it has been found recently that many samples in the Debatepedia dataset are not actually query oriented \cite{laskar2022domain}. Moreover, it was also observed that fine-tuning pre-trained neural models in this dataset without considering the query incorporation could achieve almost similar performance as the query-focused summarization models \cite{laskar2022domain}. Thus, there remains a scarcity of datasets specifically tailored for creating condensed summaries of argumentative texts that are relevant to a single query.

To address the above issue, in this work, we seek to clean the Debatepedia dataset to make it usable for query-focused single document abstractive summarization of argumentative text. For that purpose, we propose a novel methodology that leverages the text generation capability of prompt-based language models \cite{liu2023prompt,instructgpt,gpt3}. To this end, we utilize ChatGPT, a powerful generative Large Language Model (LLM) developed by OpenAI\footnote{\url{https://openai.com/}} which has received a lot of attention recently due to its impressive language generation capability -- ensuring high fluency, coherence, and grammatical correctness on its generated texts \cite{qin2023chatgpt}. ChatGPT like such Generative LLMs \cite{scao2022bloom,ul2,lambda,switch_transformers,chincila,glm120,chowdhery2022palm,sanh2021multitaskt0} that leverage the prompt-based learning mechanism have obtained impressive performance in few-shot and zero-shot learning scenarios, inspiring researchers to also explore some new applications of these models, such as data annotation \cite{wang2021gptannotator,ding2022gptannotator}. In this paper, we also harness the text generation power of ChatGPT to fix the queries in the Debatepedia dataset to construct a cleaned version of the dataset that could be used for query-focused abstractive summarization of argumentative text. With extensive experiments, we validate that our proposed cleaned version of the Debatepedia dataset overcomes the limitations of the existing noisy version of this dataset.

\section{Debatepedia Dataset Limitations}

Debatepedia is a publicly available dataset of arguments and counter-arguments on debate topics, proposed by Nema et al. \cite{nema}. It contains 13,573 query-document-summary pairs. The average number of words per document, summary, and query in the Debatepedia dataset is 66.4, 11.16, and 9.97, respectively. The dataset covers a wide range of topics, such as politics, sports, and technology, and has been extensively used in recent years to build query-based summarization models for argumentative text. 

However, the quality of Debatepedia as a dataset for query-based summarization has lots of limitations (see Table \ref{tab:q_e_1} for some examples), as it has been found recently that many queries in this dataset are not relevant to the document \cite{laskar2022domain}. Based on a randomly sampled 100 instances, it has been found in a recent study \cite{laskar2022domain} that:

\begin{itemize}
    \item 52\% of the queries in this dataset have no relevance to the documents or the summaries, as demonstrated in Table \ref{tab:q_e_1}.
    \item 70\% of the queries are close-ended (i.e., Yes/No type) questions (see Example 4 in Table \ref{tab:q_e_1}). 
    \item Though, many queries in this dataset are relevant to the documents but the summaries are more of generic due to shorter document length. Note that the average size of the document in this dataset is only 66.4 words on average.
\end{itemize}

In addition, many instances in this dataset only contain one word summary (see Example 2 in Table \ref{tab:q_e_1}) for a given query that appears both in the training and evaluation sets, which may also help the model to memorize such words for similar queries during the training phase. These issues may lead to an unexpected increase in the ROUGE score when the model starts learning to reproduce those words in the summary during the evaluation phase. Furthermore, we also find some instances where the length of the summary is longer than the document length, which usually happens in short documents (see Example 3 in Table \ref{tab:q_e_1}). 


To address these limitations, we propose a methodology for cleaning the Debatepedia dataset via leveraging ChatGPT as the data annotator to regenerate the queries. In the following, we describe our methodology.

\section{Our Annotation Methodology}

The recently released ChatGPT model has demonstrated impressive performance to solve a wide-range of problems, from generating fluent and coherent summaries from documents to solving mathematical problems, along with solving challenging information retrieval tasks, such as open-domain question answering, neural machine translation, writing programming solutions, and etc \cite{qin2023chatgpt, guo2023close}. In this work, we leverage ChatGPT as the annotator to fix the issues in the Debatepedia dataset to use it for query-focused abstractive summarization. We denote our \textbf{C}hatGPT annotated cleaned dataset for \textbf{Q}uery Focused Abstractive \textbf{Sum}marization based on \textbf{D}ebate\textbf{p}edia as the \textbf{CQSumDP} dataset.

\begin{figure*}[h]
    \centering
    \includegraphics[width=0.8\textwidth]{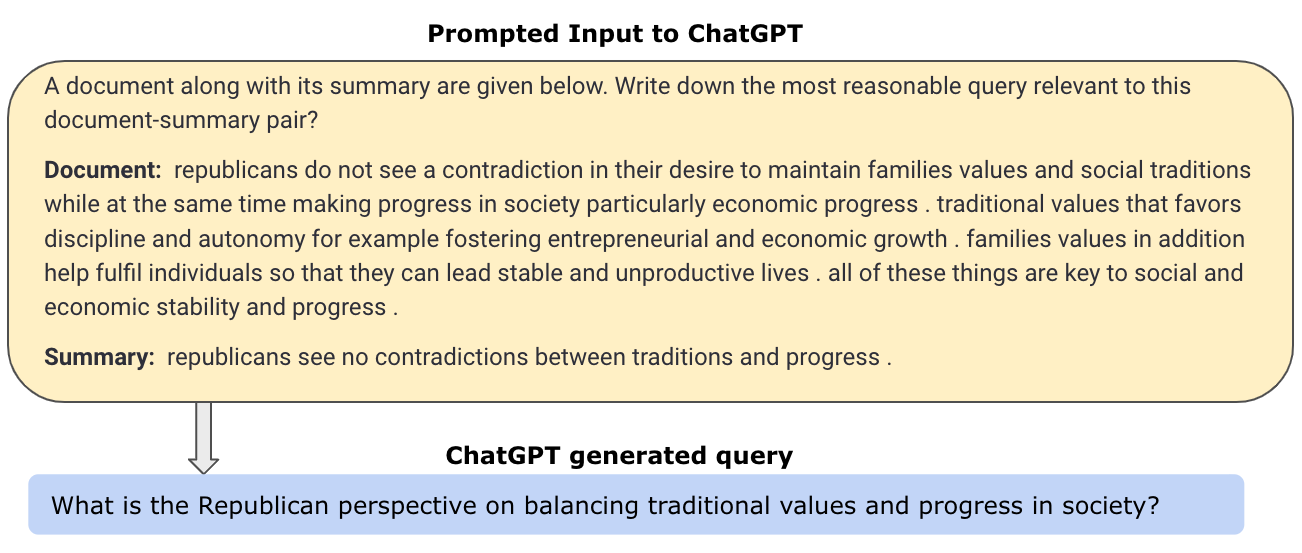}
    \caption{Our Input Prompt to ChatGPT for Query Generation}
    \label{fig:annotation_overview}

\end{figure*}

\begin{table*}
\small
\centering

\begin{tabular}{ccccc}
\toprule
\textbf{Split} & \textbf{Total Number of Samples} &  \textbf{Avg. Query Length} & \textbf{Avg. Document Length}  & \textbf{Avg. Summary Length} \\  \midrule
Training & 5212 & 11.64 & 106.82  & 9.77  \\ \midrule
Validation & 301 & 11.54 & 107.22  & 9.62 \\ \midrule
Test & 401 & 11.90 & 104.75  & 9.77 \\ 
\bottomrule
\end{tabular}
\caption{Data distribution on each split (train/valid/test) in our cleaned annotated version of Debatepedia: The CQSumDP Dataset.}
\label{table:dataset_stat}
\end{table*}

As demonstrated in the previous section the Debatepedia dataset has several limitations, containing noisy and irrelevant contents (e.g., queries/documents/summaries). To address these issues, we first clean the Debatepedia dataset to sample relevant instances from the dataset. Our objective for data sampling here is that the selected samples in the dataset could then be more relevant for query-focused summarization. Afterward, the sampled instances are used for data annotation using ChatGPT. Below we first describe our data sampling technique, followed by our approach of using ChatGPT as the annotator to construct the CQSumDP dataset.

\subsection{Cleaned Data Sampling}

Our data sampling strategy to use a cleaned version of the dataset for query focused abstractive summarization is as follows:

\begin{itemize}
    \item 

We set a minimum threshold of 75 words for the length of each selected document. This is because for the smaller-sized documents, the reference summaries are mainly the overall generic summary of the document where the additional query does not help. By excluding these smaller-sized documents by using a threshold, we can ensure that the reference summaries are more query-focused. Furthermore, setting the threshold at 75 words also helps us to address the noisy scenario in the Debatepedia dataset when the reference summary length is longer than the document length.
    
    \item As we demonstrated in Section 3 that many summaries in the Debatepedia dataset are very short (there are many summaries of only 1 word length too), we exclude instances where the length of the summary is shorter than 5 words. This helps us to clean the dataset in a way such that instead of having a dataset with very short answers, we rather propose a dataset consisting of concise but coherent and fluent summaries. This helps us to keep the dataset more relevant to summarization instead of close-ended question answering.

\end{itemize}

\subsection{Using ChatGPT for Data Annotation}

As ChatGPT like LLMs has the impressive capability to solve tasks based on the given prompt \cite{qin2023chatgpt,guo2023close}, we manually construct a prompting template that asks ChatGPT to generate the query for a given document-summary pair. Prompt learning is a technique where a machine learning model is trained to complete a task based on the prompted input \cite{liu2023prompt,sanh2021multitask}. This approach involves presenting the model with a prompt (i.e., a partial input), and the model is then tasked with generating the complete output. Prompt learning has become increasingly popular due to its ability to generate highly accurate results with very little data. It is also highly flexible, as it allows the user to modify the prompt to achieve the desired result. We show an example prompt in Figure \ref{fig:annotation_overview} where ChatGPT is asked to generate a query that is relevant to the given document-summary pair. 

 The ChatGPT version that we used for data annotation was based on the version that was last released\footnote{\url{https://help.openai.com/en/articles/6825453-chatgpt-release-notes}} on January 30th. We choose ChatGPT over other text generation models due to its impressive capability of generating high quality responses \cite{qin2023chatgpt,guo2023close} while being free to use (in contrast to their powerful models in OpenAI that requires the use of paid API subscription). One of the key reasons for ChatGPT to generate human like responses is because it was trained using the reinforcement learning from human feedback technique \cite{qin2023chatgpt,guo2023close,instructgpt}. In this technique, the model generates a response to a user's input, and then humans provide feedback on the quality and appropriateness of the response. This helps the model to generate human like responses while ensuring high accuracy, appropriateness, and fluency. For these reasons, we use ChatGPT for data annotation. 

 Though prior research has demonstrated that many queries in the Debatepedia dataset have no relevance with the document \cite{laskar2022domain}, there does not have any major issues found on the summaries in the Debatepedia dataset. Thus, we use both the document and the summary as input to ChatGPT since we already cleaned the Debatepedia dataset by removing noisy instances where the summary length is very small or exceeds the document length. While we could ask ChatGPT to generate a query followed by a query-based summary by only giving the document with the input prompt, we did not do so as it has been observed that ChatGPT tends to generate longer summaries \cite{qin2023chatgpt} and so we use both the document and the summary as input to only regenerate the queries in the Debatepedia dataset. This also allows us to use the original gold reference summaries in our proposed CQSumDP dataset without any modification.

\begin{table*}[h]
\small
\centering
\begin{tabular}{ccccc}
\toprule
\textbf{Model} &\textbf{Dataset} &  \textbf{ROUGE 1} & \textbf{ROUGE 2} & \textbf{ROUGE L} \\
\midrule
BART-Base & CQSumDP &42.26 & 22.45 & 38.84 \\ 
Pegasus-Base & CQSumDP &36.01 & 16.30 & 32.59 \\ 
T5-Base & CQSumDP &39.95 & 21.24 & 36.79 \\ 
\midrule
\addlinespace
BART-Base & Original Debatepedia & 39.97 & 21.50 & 36.87 \\ 
Pegasus-Base & Original Debatepedia & 29.70 & 11.91 & 26.77 \\ 
T5-Base & Original Debatepedia& 37.68 & 18.92 & 34.49 \\ 
\bottomrule
\end{tabular}
\caption{Performance of different models trained and evaluated on the respective versions of the Debatepedia dataset. }
\label{tab:rouge-scores}
\end{table*}

\begin{table*}[h]
\small
\centering
\begin{tabular}{cccccc}
\toprule
 \textbf{Model} & \textbf{Training Dataset} & \textbf{Evaluation Dataset} &\textbf{ROUGE 1} & \textbf{ROUGE 2} & \textbf{ROUGE L} \\
\midrule
BART-Base & CQSumDP & MS-MARCO & 44.01	&26.95&	38.34 \\
Pegasus-Base & CQSumDP & MS-MARCO & 50.34 & 33.07 & 45.80 \\
T5-Base & CQSumDP & MS-MARCO & 48.90 & 28.66 & 43.84  \\  \midrule
BART-Base & Original Debatepedia & MS-MARCO & 43.09	&23.72	&37.90 	\\
Pegasus-Base & Original Debatepedia & MS-MARCO & 46.94 & 29.24 & 42.42 \\
 T5-Base & Original Debatepedia& MS-MARCO &47.85 & 27.89 & 42.81 \\ \midrule
BART-Base & MS-MARCO & CQSumDP & 28.42 &	10.30 &	23.56  \\ 
BART-Base  & MS-MARCO & Original Debatepedia & 23.56 &	7.38 &	20.88   \\ 
\bottomrule
\end{tabular}
\caption{Domain generalization performance of different models trained on respective versions (CQSumDP and Original) of the Debatepedia dataset and evaluated on the MS-MARCO dataset, as well as trained on MS-MARCO and evaluated on the CQSumDP and Original versions of the Debatepedia dataset. }
\label{tab:domain generalization}
\end{table*}

A total of 5914 samples were annotated using ChatGPT. After the data annotation is completed, we create the training, validation, and test set based on the split provided by Nema et al. \cite{nema} for the original version of the Debatepedia dataset\footnote{\url{https://github.com/PrekshaNema25/DiverstiyBasedAttentionMechanism/tree/master/data}}. As we construct a cleaned version of the dataset by excluding noisy instances, the number of samples in each split in our cleaned version of the dataset is smaller than the original one. The overall statistics of our cleaned, annotated version of the Debatepedia dataset: the CQSumDP dataset is shown in Table \ref{table:dataset_stat}.

\section{Experimental Settings}

In this section, we present our experimental settings. Below, we first describe the models we use to evaluate our ChatGPT annotated cleaned version of the Debatepedia dataset, the CQSumDP dataset, followed by our model implementation details. To keep the experimental comparisons fair, we only use the cleaned samples of both versions of the dataset (e.g., 5914 cleaned samples, with 5212, 301, 401 instances in the training, validation, and test sets respectively, as demonstrated in Section 4). From now on, we refer to the version of the Debatepedia dataset that has the original queries but only contains our sampled 5914 instances as \textit{Original Debatepedia}.

\subsection{Models}
To evaluate the effectiveness of our ChatGPT annotated CQSumDP dataset, we fine-tune some state-of-the-art pre-trained sequence to sequence models \cite{bart, t5, zhang2019pegasus, goodwin2020flightpegasus}. For this purpose, we concatenate the query with the document and give as input to these models to generate the query-focused abstractive summaries as this approach has shown impressive performance in the query-focused abstractive summarization task recently \cite{laskar2022domain}. We describe these models below:

\paragraph{\textbf{BART (Bidirectional and Auto-Regressive Transformer):}} BART  \cite{bart} is a pre-trained sequence-to-sequence model based on the encoder-decoder architecture that was pre-trained on a large amount of diverse text data using the denoising auto-encoding technique to recover the original form of a corrupted document. The pre-training involved various objectives such as rotating the document, permuting sentences, infilling text, masking tokens, and deleting tokens. We use the pre-trained BART model since fine-tuning this model was found to be very effective on a wide range of language generation tasks, including abstractive summarization.

\paragraph{\textbf{T5 (Text-to-Text Transfer Transformer):}} The T5 model \cite{t5} is a transformer-based model that uses the BERT architecture. Unlike traditional BERT-based models that classify input text into a specific category, the T5 model treats all tasks such as text classification, question answering, neural machine translation, and text summarization as a sequence-to-sequence problem using various pre-training objectives. After pre-training, the model is fine-tuned to generate the output for a given input sequence in the required task, leading to impressive performance gain on many downstream summarization datasets. 

\paragraph{\textbf{Pegasus (Pre-training with Extracted Gap-sentences for Abstractive Summarization):}} Pegasus \cite{zhang2019pegasus} is a transformer-based pre-trained encoder-decoder model for abstractive summarization. Its pre-training objective involves generating summary like text from an input document. To achieve this, the PEGASUS model first selects and masks some sentences from the input document(s). It then concatenates these selected sentences to create a pseudo-summary. The model uses different approaches to select these sentences, such as randomly selecting a certain number of sentences, selecting the first few sentences, or computing the ROUGE-1 score between each sentence and the rest of the document to choose the top-scoring sentences. This pseudo-summary is then used for self-supervised learning. By pre-training on large datasets using this approach, the model achieves impressive fine-tuning performance on downstream summarization datasets.

\subsection{Implementation}

We use the HuggingFace\footnote{\url{https://huggingface.co/}} \cite{Wolf2019HuggingFacesTS}library to implement the baseline models for performance evaluation. Similar to the prior work, we concatenated the query with the document to give as input to the pre-trained baselines (i.e., BART, Pegasus, T5). The pre-trained model is then fine-tuned using $4$ NVIDIA V100 GPUs. The training batch size for BART was set to $16$, while  it was set to $4$ for Pegasus and T5. The other hyperparameters were similar for all models, with the learning rate being set to $2e-3$ and the maximum input (i.e., the concatenated query and document) sequence length being $150$ tokens. The minimum and the maximum target (i.e., the generated summary) sequence lengths were $5$ and $25$, respectively. A total of $10$ epochs were run to fine-tune the pre-trained summarization models. We computed the ROUGE \cite{rouge} scores in terms of ROUGE-1, ROUGE-2, and ROUGE-L using the \textit{Evaluate}\footnote{\url{https://huggingface.co/spaces/evaluate-metric/rouge}} library to compare the performance of different models on the respective test set.

\section{Results \& Discussions}

We conduct a series of experiments to evaluate the performance of strong baseline models in our proposed cleaned annotated version of Debatepedia: the CQSumDP dataset. In this section, we present our experimental findings. 

\subsection{Effectiveness of ChatGPT Generated Queries}

\begin{table*}[h]
\small
\centering
\begin{tabular}{cccc}
\toprule
\textbf{Model} & \textbf{ROUGE-1} & \textbf{ROUGE-2} & \textbf{ROUGE-L} \\
\midrule
BART-Large & 51.66 & 33.96 & 49.03 \\ \midrule
BART-Base & 42.26 & 22.45 & 38.84 \\
\bottomrule
\end{tabular}
\caption{Performance comparisons based on model size between BART-Large and BART-Base on CQSumDP.}
\label{tab:model_scaling}
\end{table*}

\begin{table*}[h]
\small
\centering
\begin{tabular}{cccc}
\toprule
\textbf{Model} & \textbf{ROUGE-1} & \textbf{ROUGE-2} & \textbf{ROUGE-L} \\
\midrule
BART-Large & 51.66 &	33.96	& 49.03 \\ 
\midrule
\textit{without query incorporation} & 46.45	& 29.92	& 44.11 \\
\bottomrule
\end{tabular}
\caption{Ablation test results after removing the query relevance in the CQSumDP dataset.}
\label{tab:ablation}
\end{table*}

\begin{table*}[h]
\small
\centering
\begin{tabular}{ccccc}
\toprule
\textbf{Model} &\textbf{Evaluation Dataset} &  \textbf{ROUGE 1} & \textbf{ROUGE 2} & \textbf{ROUGE L} \\
\midrule
Pre-trained BART-Large & CQSumDP & 26.86 &	9.46 &	21.70 \\ 
\midrule
\addlinespace
Pre-trained BART-Large & Original Debatepedia & 21.60	&6.04	&18.52 \\ 
\bottomrule
\end{tabular}
\caption{Zero-Shot Learning Performance of different models on the respective evaluation sets of Debatepedia. }
\label{tab:zeroshot}
\end{table*}

\begin{table*}[h]
\centering
\tiny
\begin{tabular}{p{0.5cm}p{2cm}p{2cm}p{6cm}p{2.5cm}}
\toprule
\textbf{\#} & \textbf{Original Query} & \textbf{ChatGPT  Query} & \textbf{Source Document} &\textbf{Gold Summary} 

 \\ \midrule

1. &  military : 

  & What actions did the government take to improve the situation for U.S. troops and veterans?

  &   provided better body armor to our troops . provided the department of veterans affairs ( va ) with more than \$ \# . \# billion to improve services to america s veterans . ended media blackout on war casualties and the return of fallen soldiers to dover afb . announced creation of a joint virtual lifetime electronic record for members of the u.s. armed forces to improve quality of medical care . ended the previous stop-loss policy that kept soldiers in iraq/afghanistan longer than their enlistment date . signed the veterans health care budget reform and transparency act authorizing advance appropriations for the department of veterans affairs by providing two-fiscal year budget authority thus enabling better medical care for veterans . endorsed by the american legion american veterans blinded veter ... ans association 

 &	 improved services benefits and respect for troops . 

\\ \midrule

2. &  we economy : has wto benefited the economy of the united states ? 

  & Has NAFTA caused job losses in the U.S? &   `` nafta and job losses '' . cyril morong ( PhD ) the wall street journal may \# \# - `` did nafta cause the u.s. to lose so many jobs [ citing figures provided in the range of \# million and \# \# ] especially high-paying manufacture jobs ? probably not . i say probably since causality in any social science ( economics included ) is difficult to prove since so many factors change so quickly in the real world . but if many high-paying manufacture jobs were lost it took many years until after nafta went into effect before they were ... but what about manufacture jobs ? we had just about \# million in \# . it actually rose to \# . \# million in \# and was at \# . \# in \#.
 
 &	  nafta has decreased the number of american job

\\ \midrule

3. &  entrepreneurs: does an mba help entrepreneurs ? 
 
& Is an MBA necessary for product managers?
  &  christopher cummings . `` is an mba necessary for product managers ? '' product management meet pop culture . february \# \# : `` hindsight . looking back the brass tacks of my mba experience were about the basics of management economics and business strategy . could that have been picked up on the job ? maybe . [ ... ] however the more important throughline of the experience relates to critical thinking perspective and learning when to lead and when to follow . [ ... ] on the job especially as a young pm it can be easy to lose perspective to miss the forest for the trees . at the time i was definitely into the plate-spinning the go-go-go the tactics and day-to-day . no time to think ; just keep moving . [ ... ] the mba experience

 & mba teach strategy plan not just tactics \\

\bottomrule
\end{tabular}
\caption{Comparisons between the original queries and the ChatGPT generated queries in some samples of the Debatepedia dataset. Note that the personally identifiable information in this dataset is anonymized with the \# token.}
\label{tab:quality}
\end{table*}

To investigate the effectiveness of our CQSumDP dataset that leverages ChatGPT to generate the queries, we compare the performance of BART, Pegasus, and T5 models on both the CQSumDP and the Original Debatepedia datasets (results are given in Table \ref{tab:rouge-scores}). We use the Base versions of these models from HuggingFace \cite{Wolf2019HuggingFacesTS}, and trained and evaluated on the respective datasets.

From Table \ref{tab:rouge-scores}, we find that all three models perform better in the CQSumDP dataset in comparison to their performance on the Debatepedia dataset. This gives a strong indication that the queries generated by ChatGPT are more helpful to improve the model performance. While comparing the performance between different models, we found that BART outperforms the other two models on both datasets in all three ROUGE metrics. More specifically, in the CQSumDP dataset, BART achieves the highest ROUGE-1 (42.26), ROUGE-2 (22.45), and ROUGE-L (38.84) scores. Though in the Original Debatepedia dataset, BART also outperforms other models by achieving ROUGE-1, 2, and L scores of 39.97, 21.50, and 36.87, respectively; its performance on the Original Debatepedia is much lower than its performance on the CQSumDP dataset. 

Our experimental results show the effectiveness of our proposed CQSumDP dataset that helps all these models to obtain better ROUGE scores than their counterparts on the Original Debatepedia dataset. The poor performance of these models on the Original Debatepedia dataset compared to the CQSumDP dataset further demonstrates the limitations in terms of query relevance in the Original Debatepedia. 

\subsection{Generalization Capability}

In the previous section, we find that all the baseline models fine-tuned on our CQSumDP dataset perform better than their counterparts that are fine-tuned on the Original Debatepedia dataset. In this section, to further study the relevance of the ChatGPT generated queries in our proposed CQSumDP dataset, we evaluate the performance based on domain generalization. In this regard, we use the similar setting of Laskar et al., \cite{laskar2022domain} where they used the QA-NLG version of the MS-MARCO dataset \cite{msmarco} to fine-tune their query-focused summarization model for abstractive answer generation and then evaluate on Debatepedia. We also use the MS-MARCO dataset for our analysis based on the following two scenarios:

\begin{itemize}
    \item \textbf{Training: MS-MARCO, Evaluation: Debatepedia:} In this scenario, we trained the baseline models on the training set of MS-MARCO (153725 samples) and evaluated on the respective versions of the Debatepedia dataset (CQSumDP and Original Debatepedia). 
    \item \textbf{Training: Debatepedia, Evaluation: MS-MARCO:} In this scenario, we do the opposite, as we trained the baseline models on the respective versions of Debatepedia and evaluated on the development set of MS-MARCO (12467 samples). 
\end{itemize}

We show our results in Table \ref{tab:domain generalization} and observe that the domain generalization performance in both scenarios: (i) while using Debatepedia for training to evaluate on MS-MARCO, as well as (ii) while using MS-MARCO as the training data for evaluation on Debatepedia, the performance is better when the CQSumDP version of the Debatepedia dataset is used in comparison to the scenario when the Original Debatepedia is used. These findings further establish the effectiveness of using ChatGPT generated queries for the query-focused summarization task in the Debatepedia dataset.

\subsection{Performance Based on Model Scaling}
So far, in our prior experiments, we utilize the Base version of each model and investigate the effectiveness of our proposed CQSumDP dataset. Though smaller models are preferred over larger models in real-world industrial scenarios where computing resources are limited \cite{laskar2022blink,laskar2022auto}, in this section, to set a benchmark performance in our proposed CQSumDP dataset, we investigate how much performance gain we can achieve via scaling to a larger model. For this purpose, we select the best performing BART model  \cite{bart} and compare its performance between its Base and Large versions in our dataset. From our experimental results given in Table \ref{tab:model_scaling}, we observe that the ROUGE score is improved by a large margin (on average an improvement of 10.37 out of all three ROUGE metrics) when the BART-Large model is used. This indicates that the utilization of the ChatGPT generated queries in the CQSumDP dataset also helps the larger summarization models to understand the query representation better, leading to an improved ROUGE score.

\subsection{Ablation Tests}

It was recently found that even without incorporating query relevance, the summarization models could achieve performance on the Debatepedia dataset almost similar to what could have been achieved via incorporating query relevance \cite{laskar2022domain}. While analyzing the Debatepedia dataset, we observe that this happens mostly on instances where the document size is quite small. As we already cleaned the Debatepedia dataset by removing such instances (e.g., short documents or summaries), in this section, we conduct ablation studies to investigate the importance of query relevance in the cleaned version of the dataset. For this purpose, we remove the query relevance while giving the input to the best performing BART-Large model and investigate the effect of removing the query in our proposed dataset. We show our results in Table \ref{tab:ablation}. We find from the table that the performance is dropped by a huge margin when the query is removed from the input text, demonstrating the importance of the query in our proposed CQSumDP dataset.

\subsection{Zero-Shot Learning Performance}

In recent times, the zero-shot evaluation of large pre-trained language models on text generation tasks, such as abstractive summarization has been on the rise \cite{gpt3,qin2023chatgpt,guo2023close}. To establish a benchmark in our proposed dataset, we also conduct a zero-shot evaluation of the best performing BART-Large model in both the CQSumDP and the Original Debatepedia datasets. To do so, we combine the query with the document and give as input to the pre-trained BART-Large model. We observe from Table \ref{tab:zeroshot} that in terms of zero-shot evaluation, the pre-trained BART-Large model evaluated on our dataset performs better than its performance on the Original Debatepedia, further establishing that the utilization of ChatGPT generated  queries in CQSumDP is more helpful than the original queries in the Debatepedia dataset.

\subsection{Qualitative Analysis of the Annotated Data}

In this section, we do some qualitative analyses between the queries in the Original Debatepedia dataset as well as the queries generated using ChatGPT in our proposed CQSumDP version of the Debatepedia dataset. For our analysis, we collect a set of 3 samples from this dataset and present them in Table \ref{tab:quality}. While comparing between the queries in the first example in the table, we find that the original query is just one word length and very ambiguous, while the ChatGPT generated query is more descriptive and more relevant to both the document and the summary. For the second example, we find that even though the original query is descriptive, it does not have any relevance to the generated summary. Whereas the ChatGPT generated query is very relevant to both the document and the summary. For the third example, we find that the original query is related to ``entrepreneurs''. However, the document is about ``product managers'', not ``entrepreneurs''. Meanwhile, the ChatGPT generated query is also very relevant to the document. This analysis further demonstrates the relevance of our ChatGPT generated query in comparison to the original query in Debatepedia.

\subsection{Cost Efficiency Analysis} 

 Recently, it was shown that using the GPT-3 \cite{gpt3} model could significantly reduce the labeling cost without sacrificing the model's performance much, making it possible to train models on larger datasets without the need for extensive manual labeling \cite{wang2021gptannotator,ding2022gptannotator}. However, to use GPT-3, it requires the use of its API\footnote{\url{https://platform.openai.com/docs/models}}, which is not free. On the contrary, ChatGPT is free to use. Meanwhile, we observe that generating the query in the Debatepedia dataset was also quite fast, as we observe that we could generate the queries for about 4 samples on average per minute while using ChatGPT for data annotation. This is also quite fast than giving the data for human annotation, as the human not only needs to read the document and the summary, but also needs some time to think about what could be the most effective query for the given document-summary pairs. Thus, in terms of both cost and time, it is more efficient to use ChatGPT for data annotation.

\section{Conclusions and Future Work}
In this paper, we presented a methodology for cleaning the Debatepedia dataset to make it suitable for query-focused abstractive summarization. We removed the noise from the dataset to construct a cleaned version of the dataset while using ChatGPT's language generation capabilities to address the limitations of the queries in this dataset. Our approach results in a cleaner version of Debatepedia that is found to be very effective for training and evaluating query-focused summarization models via outperforming the original dataset in terms of query relevance and summary generation quality. This indicates that our cleaning approach is effective in improving the dataset's quality for research in summarization. 

In the future, we will explore if the chain of thought prompts \cite{wei2022chainofthought} with ChatGPT leads to better query generation. We will also explore the performance of fine-tuning other pre-trained models on our proposed dataset \cite{t0,BLOOMZ,chowdhery2022palm}. In addition, we will investigate the potential of using ChatGPT as the annotator for other tasks in Information Retrieval \cite{lin2021pretrained,laskar-LREC,laskar-etal-2022-blink,laskar2022auto,JH1,JH4,JH5} to assess its generalizability. Finally, we will release our annotated version of the Debatepedia: the proposed CQSumDP dataset to encourage further research in the query-focused abstractive summarization task.

\section*{Acknowledgements}

We would like to thank OpenAI for making ChatGPT freely available which helps us to use it for data annotation. 
This research is supported in part 
by the Natural Sciences and Engineering Research Council (NSERC) of Canada 
and the York Research Chairs (YRC) program.

\bibliography{anthology,custom}
\bibliographystyle{acl_natbib}

\end{document}